# Font Acknowledgment and Character Extraction of Digital and Scanned Images

Syed Muhammad Arsalan Bashir
Institute of Space Technology
1, Islamabad Highway
Islamabad, Pakistan

## ABSTRACT
The font recognition and character extraction is of immense importance as these are many scenarios where data are in such a form, which cannot be processed like in image form or as a hard copy. So the procedure developed in this paper is basically related to identifying the font (Times New Roman, Arial and Comic Sans MS) and afterwards recovering the text using simple correlation based method where the binary templates are correlated to the input image text characters. All of this extraction is done in the presence of a little noise as images may have noisy patterns due to photocopying. The significance of this method exists in extraction of data from various monitoring (Surveillance) camera footages or even more. The method is developed on Matlab© which takes input image and recovers text and font information from it in a text file.

## 1. INTRODUCTION
Font recognition has been a subject of studies for many decades as it is of great importance. For the online databases all the documents, letters, etc. must be in a digital text form. So to word process this data. It was never so easy to identify any text since there are thousands of fonts having difference in the letters shape. So a system which should be efficient enough to identify them with great amount of accuracy is always a need. This paper is based upon one of the easiest and efficient recognition scheme where there are reference fonts characters stored in the form of binary image matrix. Then the given image is read and split up into lines and characters. Then every single letter detected is compared with the database letters and after detecting the resemblance the output is generated having font info and the output text that was in the text.

Now every text image has its unique size of information so the scheme uses resizing technique so that the detected letters are of the same size as that of the templates (reference letters). Firstly the image is converted from a colorful version to BW (black and White) image. Which is then scanned for letters every shape having less that 25 pixel is ignored, and this is the smallest size of text that can be recovered.

There are three types of font recognition namely: Mono-font, Multi-font, and Omni-font. Mono-font systems are those which deals with the single font document and hence are very accurate. In Multi-font there are several fonts which resembles to all fonts are used as references so they are useful in documents having many fonts in a single page but since not all the font are present so the accuracy is compromised. Finally Omni-font system detects all the fonts using a single font as a reference usually Arial is used as it resembles to all other fonts in a way that error is very small.

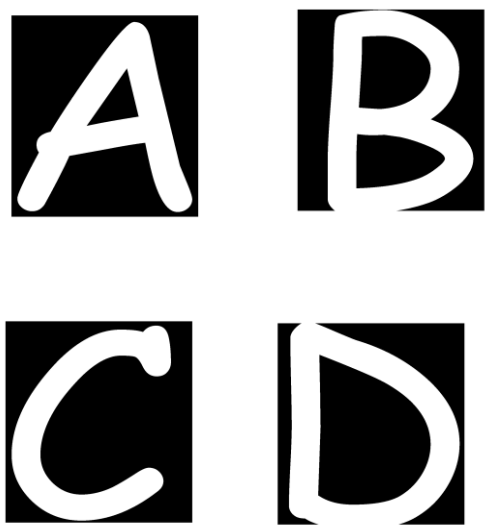

**Figure 1: Reference font Comic Sans MS**

Several models are already present like Arabic font recognition in [1] uses Hidden Markov Model (HMM) which was previously used for speech recognition. Likewise [2] uses the ApOFIS approach, in which it is assumed that one the text is of the same font and is homogeneously placed.

This paper deals with a Multi-font system which has a subset of three fonts namely: Times New Roman, Arial and Comic Sans MS (see Figure 1). Using these three fonts the images are converted to characters along with their font information. Only capital letters are implemented but the small letters can also be detected if the templates are made using the template building code. In [3] templates based Arabic character recognition method was explained in which font recognition was done by segmentation text into symbols and comparing with training symbols already there. From [4-8] different systems were used for different form of scriptures like Oriya and Chinese.

In [9] a technique was used in which only few basic function words (the, of, and, a and to) were used to identify the font. In [10] scale-invariant feature transform (SIFT) is used to extract features based on training fonts and images with high accuracy of 93.37% for Thai fonts. In [11] an analytical approach is used for segmenting Arabic font and extracting features.

## 2. DESIGNING OF TEMPLATES

The entire reference template used in this design is in Black and White form, the total size of the letters generated is 46pixel x 26pixel and all the letters of the fonts are then input using command:

$$ca = imread('font\textbackslash ca.bmp'); \qquad (1)$$

This command will store bitmap image 'ca.bmp' in 'ca' variable in the form of a logical binary 46row x 26column matrix. All the characters are then similarly stored and converted to their respective binary matrix. There is requirement of a single array containing all font characters so that all the letters may be compared using a single for loop otherwise long coding would have been required to correlate such huge amount of characters. This procedure generates a 46pixel x 26pixel binary matrix (see Figure 2).

By using multi-dimensional array the binary information is saved so to form a set of reference character array. This is done by using mat2cell function which converts the multi-dimensional matrix to a cell with same information stored.

So an array is formed by using 'mat2cell( )' function:

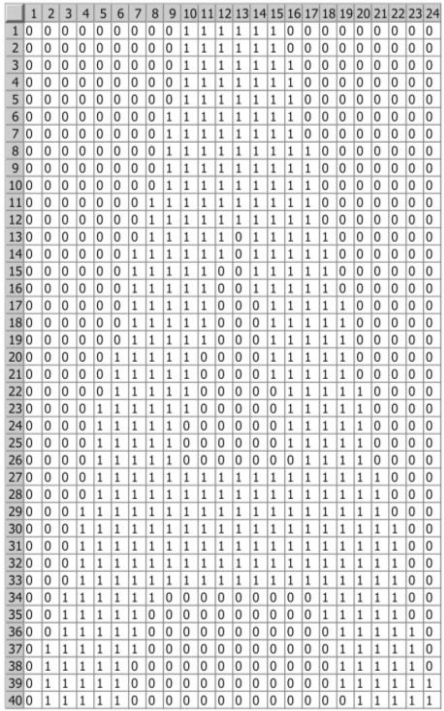

**Figure 2: Template for Arial character 'A'.**

templates_font = mat2cell(character,[46],[26

26 26 26 26 26 26 26 26 26 ...

26 26 26 26 26 26 26 26 26 ...

26 26 26 26 26 26 26 26 26 ...

26 26 26 26 26 26 26 26 26 ...

26 26 26 26 26 26 26 26 26 ...

26 26 26 26 26 26 26 26 26 ...

26 26 26 26 26 26 26 26 26 ...

26 26 26 26 26 26 26 26 26 ...

26 26 26 26 26 26 26 26 26 ...

$$26\ 26\ 26\ 26\ 26\ 26\ 26\ 26]); \qquad (2)$$

Code in (2) generates a single array having 1row x 108column and each cell having a 46row x 26column. Then this array is saved in the form of a template so it is called in the main program for correlation.

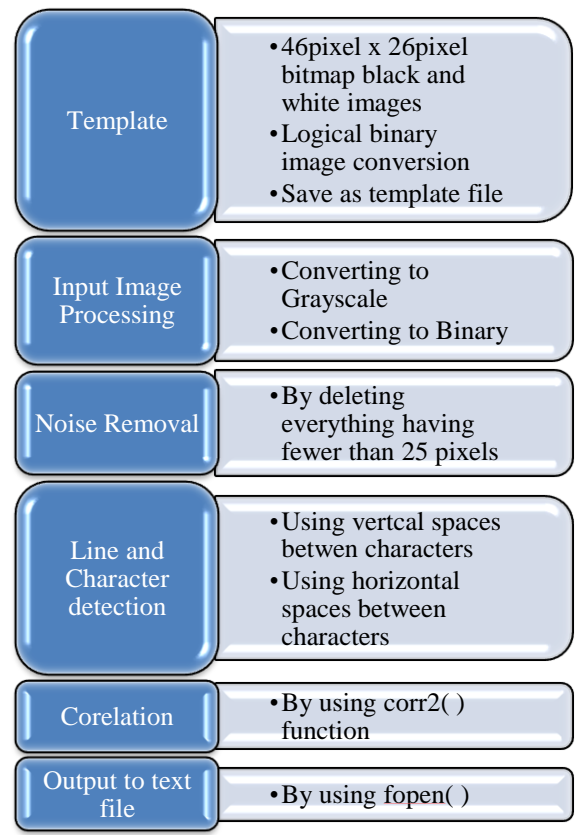

**Figure 3: Block diagram of whole system.**

## 3. FONT RECOGNITION & RECOVERY

The whole system consists of six steps (see Figure 3) but based upon the function the process is divided up into four basic parts.

### 3.1 Character detection and Image resizing

This process starts with the processing of the input image and converting it to such a form which is desired i.e. binary format. Usually the input file is color image, so if it is RBG (color) image then it is made gray scale image using command:

$$image = rgb2gray(imagen); \qquad (3)$$

This command converts the input image 'imagen' to gray scale and saves it in 'image'. Then binary image is generated using command:

$$\text{im2bw(imagen,threshold);} \qquad (4)$$

This command makes a binary image using '1' for White color and '0' for Black color hence binary image matrix is generated.

To separate rows and columns spacing is used. All the details having fewer than 25pixels are ignored hence if there is noise which appears when any document is scanned or photo-copied will be discarded. Now rows are detected by using the principle that vertical gap appears i.e. whenever 25pixel or more gaps appears vertically there is a line change, and characters in line are detected by checking that horizontally whenever there appear a gap of 25pixel or more, there is a start of new character.

In [12] fusion method was used to enhance the efficiency in feature extraction by using combination of ring and sector method.

After detecting the characters all the characters are resized to 46rows x 26colums using:

$$\text{chr = imresize(img,[46 26]);} \qquad (5)$$

This command resizes 'img' to 46rows x 26 columns and saves it in 'chr'.

## 3.2 Image Processing
A function is called which calculates the 2-D correlation between the input character 'chr' and the reference characters stored in 'template' array.

$$\text{corr = corr2(template\{1,n\},chr);} \qquad (6)$$

2-D correlation coefficient between two 2-D arrays A and B is calculated using:

$$r = \frac{\Sigma_m \Sigma_n (A_{mn}-A^*)(B_{mn}-B^*)}{\sqrt{\{\Sigma_m \Sigma_n (A_{mn}-A^*)^2\}\{\Sigma m \Sigma n (B_{mn}-B^*)^2\}}} \qquad (7)$$

Where

$A^*$ and $B^*$ are mean2 of A and B respectively

## 3.3 Error Correction
This algorithm calculates the correlation coefficient of all the separated characters and selects the maximum value out of it. If the coefficient value is less than a preset value which in this case was +0.4 then it discards the detected character and displays an error. The algorithm consists of:

- For (every character (chr ∈ chr_line)) {Scan characters in template array 'template( )' Select template(j) ∈ template such that corr2(template{1,n},chr ) is the maximum obtained }.

- Detect if max correlation coefficient detected is less than +0.4, if yes then discard the character.

- For each error detected add a full stop '.' in the text file to indicated error correction

By doing so more noise rejection was possible and data recovery was possible even in relatively big noisy samples in the image

## 3.4 Font Detection
After detecting the character a counter is increased to indicate the font template it belonged to and the number of correlated template characters is used to regenerate the detected font using the following code:

if (arial>comic)&&(arial>times)

fprintf(fid,'%s\n','The font is Arial and the text is:');  (8)

Code in (8) compares the counter 'arial' with 'comic' and 'times' and prints the output in a text file using text write code in Matlab.

fid = fopen('DSP.txt', 'wt');

fprintf(fid,'%s\n\n',word);  (9)

Hence the text is detected and is saved in a text file named 'DSP.txt'.

In order to remove noise which is of larger pixel size the condition is implemented that if correlation is less than 0.4 ignore the character as it is noise. By doing so more noise rejection was possible and data recovery was possible even in relatively big noisy samples in the image.

## 4. SIMULATIONS AND RESULTS
A simple image (see Figure 4a) is taken along with all the reference fonts with noise and is simulated in MATLAB® [13] using source code. The input image carrying salt & pepper, gaussian and speckle noise (see Figure 4b) is converted into binary image (see Figure 4c) and then noise is removed from the image before further processing (see Figure 4d). The achieved result, along with the font information and recovered text is also shown (see Figure 5).

The procedure is performed with three different font sizes for three different fonts and the results are shown in tabular form (see Table 1). Column 'D' shows total number of correct detections in each case while column 'S %' gives success probability in percent. As it is evident the success rate increases as the font size is increased and after filtering there

**Table 1. Results of font recognition and recovery**

| Font | | Total Characters | Without filtering | | Using optimized filtering | |
|---|---|---|---|---|---|---|
| Type | Size | | D | S % | D | S % |
| Arial | 6 | 3250 | 2354 | 72.43 | 3151 | 96.95 |
| | 8 | 3100 | 2289 | 73.84 | 3049 | 98.35 |
| | 10 | 3300 | 2587 | 78.39 | 3261 | 98.82 |
| Comic Sans MS | 6 | 3250 | 2309 | 71.05 | 3134 | 96.43 |
| | 8 | 310 | 2221 | 71.65 | 3033 | 97.84 |
| | 10 | 3300 | 2457 | 74.45 | 3247 | 98.39 |
| Times New Roman | 6 | 3250 | 2388 | 73.48 | 3165 | 97.38 |
| | 8 | 3100 | 2361 | 76.16 | 3057 | 98.61 |
| | 10 | 3300 | 2782 | 84.30 | 3276 | 99.27 |

is significant improvement in success rate. The overall average success rate for the proposed system is found to be 98.01%

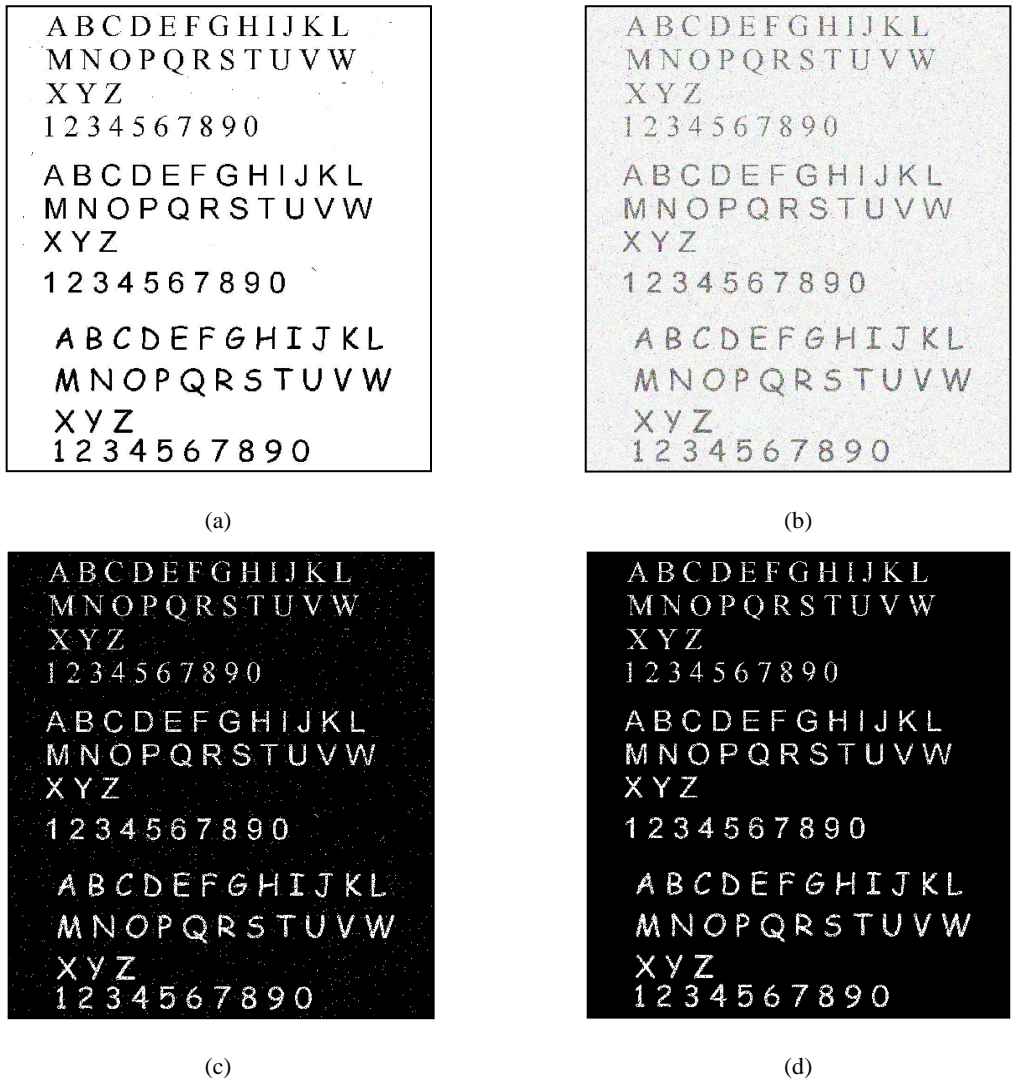

Figure 4: Input image – Noise removal

Using statistical features and geometric moment features as in [14] even higher recognition rate can be achieved by using

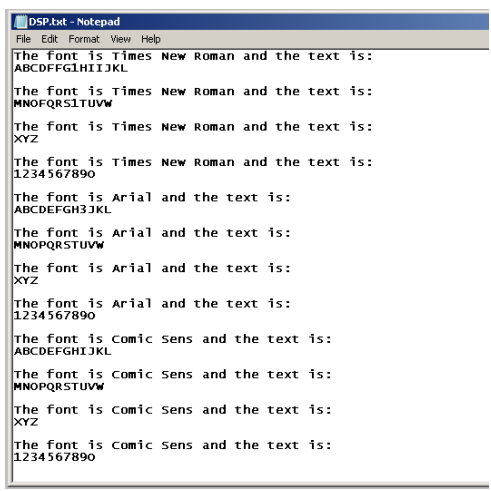

Figure 5: Recovered characters with font information

combination of fonts to detect a feature.

## 5. APPLICATIONS

Applications are huge including the making of digital copies of the ancient scripture to everyday life bills etc. that may be required to be of digital form. This setup can be used to recover textual information from surveillance footage as in [15] a hybrid approach is used to extract textual information form a video scene. Can be used in license plate recognition as in [16] an approach to determine license plate number is explained.

All the hardcopies which are of greater use as there are no softcopies of that particular script may be scanned and converted to a digital copy for re-printing purpose. In industry its use in making digital copies of all the bills and tenders so that they may be stored in a digital form i.e. in a computer database. This will save time and human effort in maintain the hardcopies and it is also more secure.

## 6. ACKNOWLEDGMENTS

Special thanks to Mr. Farhan Ali Khan Ghouri (CEO AppXone) for the support during this project. Thanks to all the colleagues who supported and encouraged since the beginning of this research based project particularly the faculty of Electrical Engineering, Institute of Space Technology, Islamabad.

I am grateful to my family as they always supported me in any way possible.